# Evaluation of Key Spatiotemporal Learners for Print Track Anomaly Classification Using Melt Pool Image Streams

Lynn Cherif[1, *], Mutahar Safdar[1, *], Guy Lamouche[2], Priti Wanjara[2], Padma Paul[3], Gentry Wood[4],
Max Zimmermann[5], Florian Hannesen[6], Yaoyao Fiona Zhao[1, **]

[1]Department of Mechanical Engineering, McGill University (e-mail: lynn.cherif@mail.mcgill.ca,
mutahar.safdar@mail.mcgill.ca, yaoyao.zhao@mcgill.ca)
[2]National Research Council Canada (e-mail: guy.lamouche@cnrc-nrc.gc.ca, priti.wanjara@cnrc-nrc.gc.ca)
[3]Braintoy AI (e-mail: ppaul@braintoy.ai)
[4]Apollo Machine & Welding Ltd (e-mail: gentry.wood@apollomachine.com)
[5]Fraunhofer Institute for Laser Technology ILT (e-mail: max.zimmermann@ilt.fraunhofer.de)
[6]BCT GmbH (e-mail: f.hannesen@bct-online.de)

*Equal Contribution, **Corresponding Author

**Abstract:** Recent applications of machine learning in metal additive manufacturing (MAM) have shown great potential to resolve critical barriers to MAM's widespread adoption. Recent research on the topic highlights the significance of using melt pool signatures to predict defects on-the-fly. While high-fidelity melt pool image data has the potential to enable accurate predictions, hardly any work exists on the use of state-of-art spatiotemporal models to leverage the information embedded in the transient and sequential nature of the additive process. This work introduces and implements some of the major deep spatiotemporal learners that can be adapted to classify melt pool image streams from different materials, systems, and applications. In this regard, two-stream networks with a spatial and temporal stream, a recurrent spatial network, and a factorized 3D convolutional neural network were investigated herein. The generalization abilities of these models to perturbations in the melt pool image data are tested using data perturbation techniques that are grounded in palpable process situations. The implemented architectures exhibit the ability to learn the spatiotemporal features of the melt pool image sequences. However, only the Kinetics400 pre-trained SlowFast network, which belongs to the two-stream category, showed strong generalisation abilities to data perturbations.

## 1. INTRODUCTION

Additive manufacturing (AM), or 3D printing, uses layer-wise material addition to fabricate parts, as opposed to subtractive manufacturing based on material removal. Offering unique benefits, AM has the potential to rival conventional manufacturing with opportunities to eliminate tooling, reduce high-value material waste, enable complex designs, support on-demand printing, simplify assembly, and allow mass customization. The use of AM to print metallic parts has gained traction in recent years. However, metal AM (MAM) remains constrained by the maturity of different technologies, particularly for high-volume production at industrial scale (Zhu et al., 2021). The challenges of MAM and their impact on the quality and performance of as-built parts are well documented (Fu et al., 2022). Solving these challenges in MAM will propel it to compete with state-of-the-art subtractive approaches in industry, and eventually realise its full potential.

Several methods exist to approach the challenges in AM. One way is through extensive experimentation to develop AM knowledge that could reduce variability in the process. Such approaches face barriers of cost, time, and effort (Johnson et al., 2020). Another technique is to model AM processes using analytical or numerical solutions. An accurate solution of this type can help understand the key process-structure-property relationships and guide AM practitioners to design and plan the process accordingly (Luo and Zhao, 2018). However, these models can be based on simplistic assumptions, which may not hold in practice and could lead to low-fidelity results. Recently, an increasing number of researchers have addressed AM challenges through empirical solutions based on data. Amongst others, these data-driven techniques offer unique advantages of computational efficiency, synergy with digital manufacturing, on-the-fly usage, physical output optimization, quality control loop closure, and knowledge transfer (Qin et al., 2022).

Machine learning (ML) has been widely researched across all lifecycle phases of MAM. ML applications at the design, process, structure, and property phases of AM are context dependent. Among these, process-oriented applications are the most frequent and are inspired by the ability of ML to provide in-line support to operators. Molten material or simply, the melt pool, is of prime importance in MAM as it is directly related to the resulting structures (macro, micro, nano) and their associated properties. ML applications in MAM employ melt pool data of different scales, spectra, fidelities, and dimensions. Among these, graphic data has been reported to be the most common method for capturing melt pool signals (Wu et al., 2021). The nature of the data inspires different handling and processing techniques of ML. In the case of graphic data, the pixel variations are learned by convolutional neural networks (CNNs). This has inspired a plethora of CNN-based applications to correlate melt pool image data with product defects (e.g., porosity, lack of fusion, cracking).

Like the initial trends in computer vision, most AM applications deploy well-known CNN architectures that use melt pool images to make structure, property, or performance predictions. Low-fidelity graphic data leads to information losses when only a 2D surface image of the 3D melt pool is captured during MAM. The problem worsens in a typical 2D CNN, which can only learn spatial correlations in the data. As a result, recent research on the topic highlights the significance of increasing the fidelity of these solutions by incorporating more information (Johnson et al., 2020). This may be realized by adding physics-based information in data-driven pipelines. However, these approaches face bottlenecks when integrated with real-world applications, such as in-situ monitoring. On the other hand, spatial feature learning by CNNs can be combined with temporal feature learning across the sequences of melt pools (Zhang et al., 2019). This approach provides a way to enhance a CNN's performance by capturing space (gradients) and time (rates) relations, representative of the complex multi-physics and multi-phase MAM processes.

This paper first explores the state of spatiotemporal learning in MAM, with a focus on sequential melt pool image data. The review is then extended to the parent domain of computer vision where the main categories of spatiotemporal architectures are presented. To increase the size of the dataset and evaluate the generalisation abilities of the models, data augmentations based on in-line and real-world applications of ML-driven MAM are applied. Representative video classifiers from each category are then selected and tested on sequential melt pool image data. The findings of these experiments are finally presented and discussed alongside potential future directions.

## 2. BACKGROUND

The application of spatiotemporal learning in MAM is relatively less widespread than pure spatial learning. The same trend holds true for melt pool data as well. A recent review on CNN-based ML applications in AM highlights the widespread adoption of learners that are based on the pixel topology of graphic data (Valizadeh and Wolff, 2022). To enable real-time melt pool control, Yang and colleagues investigated the potential of ML to characterise in-situ melt pool images; four different defect types of the melt pool were used for classification in a CNN architecture (Yang et al., 2019). In another application, melt pool images were used in a CNN to predict porosity with the aim of automating the process to characterise MAM-built parts. Zhang and colleagues found that CNNs were able to outperform other shallow models when classifying powder bed fusion additive processes based on melt pool images (Zhang et al., 2018). However, ML applications utilising CNNs for melt pool classification in MAM inherently lack the ability to capture the transient and volatile nature over the sequential process, which is embedded in the time domain. This has inspired several works to explore methods of incorporating the temporal aspects.

Recent research in MAM has seen a growing trend to utilise high-fidelity data, with space-time learning as one of its applications. Several researchers have pointed to enriching features in data-driven solutions, which can lead to high performance for the task at hand. Zhang and colleagues showed that leveraging both spatial and temporal aspects of the melt pool image data through a "hybrid" CNN (e.g., a first CNN for spatial modeling and a second CNN for temporal modeling) can achieve higher performance (99.70% > 93.50% accuracy) compared to using purely spatial CNNs (Zhang et al., 2019). Apart from melt pool monitoring and classification, several research studies in MAM have explored the potential of spatiotemporal learning; these applications constitute a diverse set, such as materials design (Yu et al., 2022), temperature profile prediction (Paul et al., 2019), and process build interactions (Yazdi et al., 2020).

While these efforts represent the growing trend of relying on space-time features in data-driven pipelines, the use of architectures that can learn these features is still limited in MAM. Video architectures are considered to be synonymous with spatiotemporal learning. A recent review on video classification grouped these methods into eight categories. The highlighted categories include hand-crafted approaches, 2D CNNs, 3D CNNs, spatiotemporal 2D CNNs, recurrent spatial networks, two/multi-stream networks, mixed convolution, and hybrid approaches (Rehman and Belhaouari, 2021). In computer vision, these spatiotemporal architectures are inspired from baseline CNNs, where the capacity to learn temporal features is added. The research focuses on developing architectures whose performance is on par with, or better than, representative video classifiers on human action recognition benchmark video datasets (e.g., UCF101, HMDB51, Kinetics400/600/700).

The simplest and most basic approach to learning temporal features is based on 2D CNNs, where features across frames are fused in different ways. This leads to architectures such as early and late fusion of features (Karpathy et al., 2014). Approaches based on two or multiple streams represent another category, where each stream of the architecture learns a different type of information (e.g., spatial and temporal features). Alternatively, the basic convolution operation that learns spatial features can be modified to extend its operation in the time dimension. These architectures can be grouped under 3D or mixed convolutional networks. A representative type of this category is a factorized spatiotemporal convolutional network (Tran et al., 2018). Finally, recurrent spatial networks that employ a sequence model in conjunction with a spatial one highlight another important category of video architectures. CNN-LSTM (long short-term memory) is the most common example (Donahue et al., 2015).

Based on the review thus far, the applications of spatiotemporal learning in MAM are limited to certain architectures (e.g., CNN-LSTM), while the aforementioned others have not been evaluated. Also, these applications did not consider the potential impact of real-world conditions that could corrupt the data in the long term (e.g., camera setup, operating conditions). Inspired by these limitations, this work evaluates state-of-the-art video architectures in the parent field of computer vision on sequential melt pool image data, and their ability to generalise to perturbations in this data using process-grounded augmentation methods.

## 3. METHODOLOGY

### 3.1 Data collection and description

The data was shared by the authors of another study (see acknowledgements for details) and the reader is referred to their work for the details of the experimental setup through which the respective melt pool images were obtained (Zhang et al., 2019). Their melt pool raw image data was collected using a high-speed camera with a sampling rate of 2000 fps during a laser powder bed fusion (LPBF) additive manufacturing process, and was used in all our experiments. Four process conditions were considered: balling (Figure 1. A), irregularity (Figure 1. B), normal (Figure 1. C), and overheating (Figure 1. D). The shared dataset contains three videos per class in the training set, and one video per class in the validation set. All videos contain 284 frames, except for one training irregularity video containing 160 frames, and a training overheating video containing 283 frames. The training data thus contains a total of 3,283 frames, and the validation set has a total of 1,136 frames.

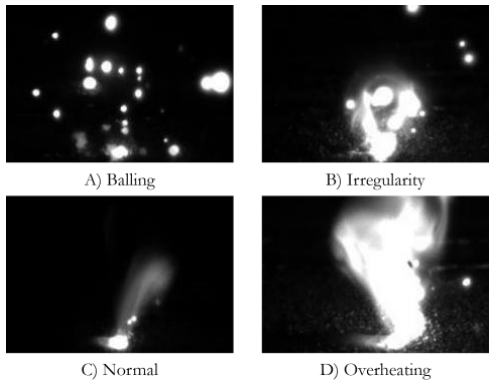

Fig. 1. Representative melt pool images corresponding to each label (Zhang et al., 2019).

The original images used are RGB of size 140 x 200 pixels. However, we resized the images using bilinear interpolation to the recommended input size of every tested classifier. Given that the track quality is determined after solidification of the melt pool, i.e., after a maximum period of $T_{max}$= 4.65 ms, we followed the recommendation of the data's authors and used a sequence of 10 frames for all networks except for SlowFast. The sequences were organized in a neighbouring window fashion. Our choice is mainly due to its higher computational efficiency compared to the sliding window approach. The labels were assigned per sequence, i.e., per clip. The input size is thus [$N$, $S$, $C$, $H$, $W$] where $N$ is the batch size, $S$ is the sequence length, $C$ is the number of image channels, and $H$ and $W$ are the image height and width, respectively. The input is reshaped to [$N$ x $S$, $C$, $H$, $W$] when initially fed to a 2D convolutional layer, and to [$N$, $C$, $S$, $H$, $W$] when fed to a 3D convolutional layer. Following the requirements of the pre-trained SlowFast model, a sequence of 64 consecutive frames was randomly sampled per video. Of the 64 frames, 32 are sampled with a temporal stride of two and fed into the fast pathway, and four are sampled with a temporal stride of 16 into the slow pathway.

### 3.2 Data augmentation

To equip the models with robustness to real-world varying process conditions and mitigate the risk of overfitting, data augmentations of the original images were conducted while preserving their time sequence. The augmentations are based on palpable in-line situations. Data augmentation has the potential to act as a regulariser by both increasing the number of training samples and introducing perturbations in the training distribution, potentially covering the distribution of the real-time process data. Figure 2 illustrates the augmentations done in this study for a sample frame of the balling category.

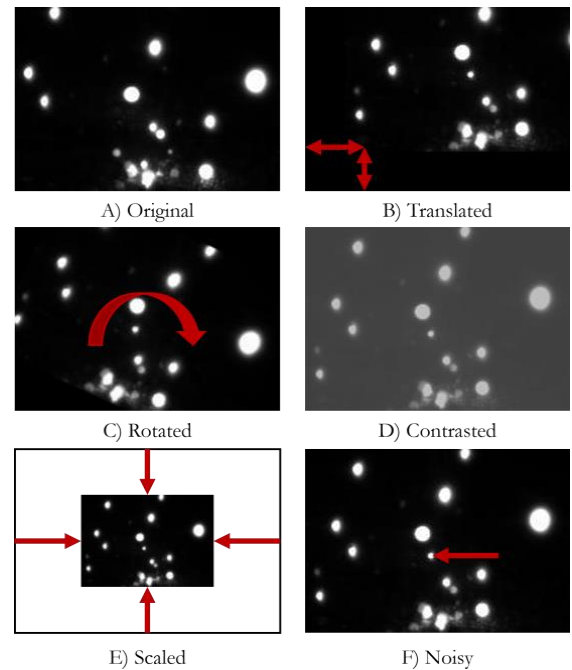

Fig. 2. An example of the data augmentations done in this study, shown for a sample frame of the balling category.

The augmentations illustrated in Figure 2 include image translations with a step size of five pixels in both the x- and y-directions {[(0,0);(+25,+25)], [(0,0);(-25,+25)]} (e.g., Figure 2. B), image rotations with steps of 5° in [-25°; 25°] (e.g., Figure 2. C), contrast adjustments with steps of 0.1 in [0;1[ (e.g., Figure 2. D), image downscaling with steps of 10% in [10%;80%] (e.g., Figure 2. E), as well as Gaussian [mean: 0, var: 0.1, std: $var^{0.5}$] and Poisson noise addition (e.g., Figure 2. F).

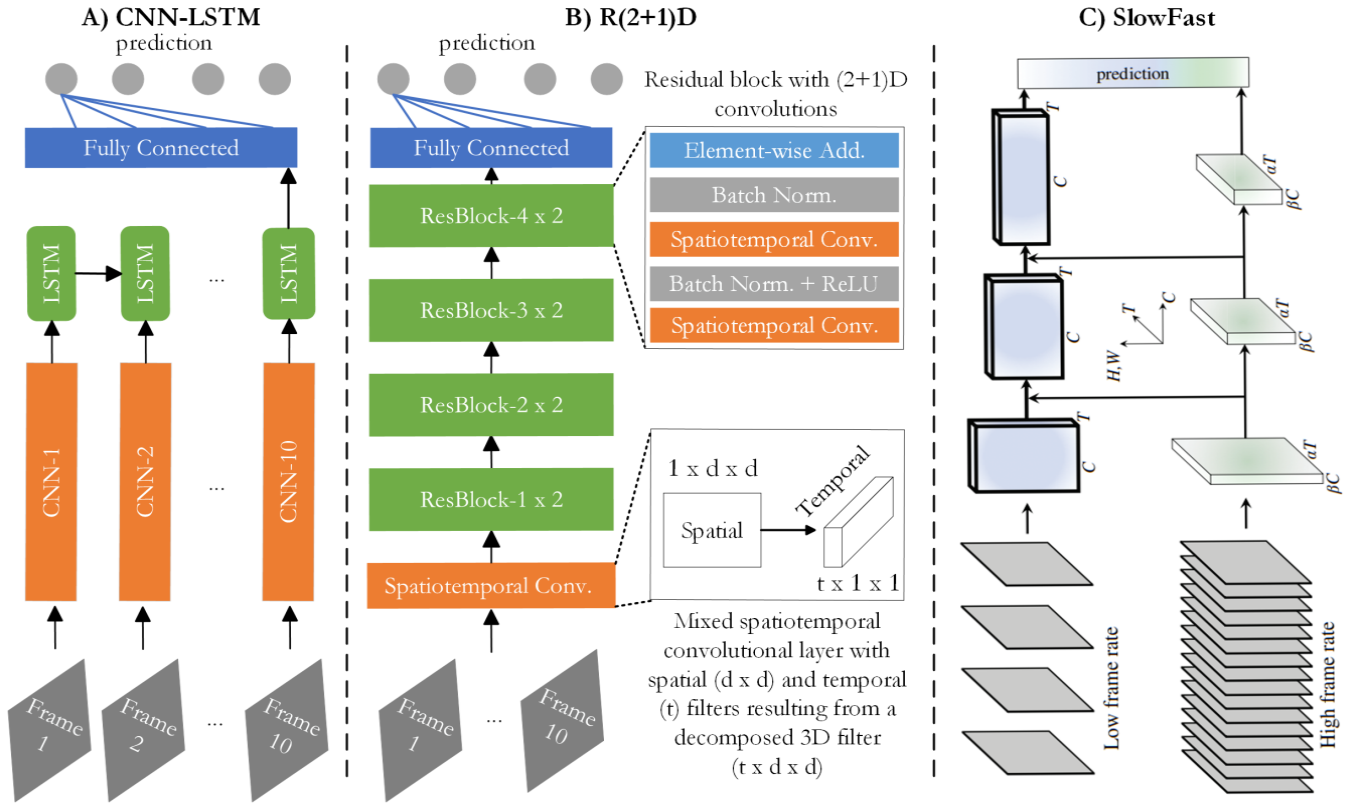

Fig. 3. Spatiotemporal architectures explored in this study. The SlowFast architecture's figure is taken from the original paper (Feichtenhofer et al., 2019)

With regards to translations, a total of five such translations were completed with a positive-x and positive-y. Another five were repeated with a negative value of x while keeping the y positive. This resulted in 10 rotations, nine contrasts, two noise additions, eight scalings, and 10 translations per datapoint, augmenting the data 39-fold. A total of 128,037 training frames and 44,304 validation frames were obtained. The final mixed dataset containing both original and augmented videos holds 131,320 training frames and 45,440 validation frames.

### 3.3 Spatiotemporal classifiers

Although shallow spatiotemporal networks demonstrated competitive performance when trained and tested on the original given dataset (Zhang et al., 2019), they do not generalise well to perturbations in the data, as will be shown in section 5. This result begs the question of whether deeper, state-of-the-art spatiotemporal classifiers can generalise better. Figure 3 shows the three candidate architectures in this regard. Given that the "hybrid" model presented by Zhang et al. (2019) is based on 2D CNNs, and that other methods based on 2D CNNs have demonstrated subpar performance compared to more recent video classifiers (Karpathy et al., 2014), two models from the multi-stream, one from the mixed convolution, and one from the spatial-recurrent categories have been selected. The choice of architectures was based on a balance between their performance, ease of implementation, training time, and data quantity requirement - ultimately eliminating pure 3D CNNs. More specifically, we selected a two-stream network that fuses a spatial and temporal stream leveraging optical flow (Simonyan and Zisserman (2014), Feichtenhofer et al., 2016), a CNN-LSTM based on long-term recurrent convolutional network (LRCN) (Donahue et al., 2015), an R(2+1)D factorised residual convolutional network (Tran et al., 2018), and a Kinetics400 pre-trained SlowFast network (Feichtenhofer et al., 2019), which is also a two-stream network.

The initial two-stream network consists of two quasi-identical "spatial" and "temporal" 2D CNNs, which are either fused at the prediction layer through average fusion, or at a convolutional layer. The temporal stream's input initially considered relies on optical flow. The reader is referred to the works of Simonyan and Zisserman (2014) and Feichtenhofer et al. (2016) for the implementation details. Dense optical flow is a set of displacement vectors computed over every pixel between a pair of consecutive frames, explicitly describing the motion between video frames. In this paper, the frames' dense optical flow was computed using the small recurrent all-pairs field transforms (small-RAFT) model (Teed and Deng, 2020). The spatial stream is fed individual frames, and the temporal stream a stack of $2L$ dense optical flow frames, where $L$ is the number of frames in an input clip. It is worth noting that the two-stream model that relies on optical flow was only considered as an interesting comparison benchmark, and would likely be computationally inefficient for online application. Due to memory requirements, issues were encountered with pre-computing optical flow using our resources. Moreover, obtaining it on-the-fly would introduce bottlenecks during training, despite small-RAFT's relatively high computational efficiency. Thus, we were unable to adequately train and evaluate the temporal stream, and hence, the two-stream model. However, efforts to benchmark the spatial stream by fine-tuning its ImageNet-pretrained version were made and are presented in section 5.

The second model, CNN-LSTM, is a type of recurrent CNN capable of learning both spatial and temporal features of the data. The particular architecture implemented in this paper is called LRCN (Donahue et al., 2015). This network is illustrated in Figure 3. A. The CNN portion of LRCN that learns spatial features is based on CaffeNet, which is a variant of the well-known AlexNet. The first fully-connected layer of CaffeNet having 4096 features is fed to an LSTM that can learn long- and short-term dependencies across frame-wise spatial features. In this case, the LSTM model only has one layer with a hidden size of 256. For each subsequence (10 frames) the last hidden state is fed to a linear layer with an output size of four, each corresponding to a label of concern.

The third model, R(2+1)D, is a type of mixed convolution that decomposes the 3D convolution operation into a spatial 2D convolution and a temporal 1D convolution. Indeed, 3D CNNs can extend the convolutional operation in the time dimension at the expense of a significantly increased number of parameters and computational cost. This has inspired researchers to investigate mixed 1D and 2D convolutions that are computationally efficient. The architecture considered in this paper is related to factorized spatiotemporal convolutional network (FSTCN). The concept can be integrated with any CNN architecture. More precisely, the architecture implemented in this paper is based on the 3D ResNet with 18 layers, where each 3D convolution layer is factorised into a (2+1)D convolution. The details of the architecture can be found in the original paper on FSTCN (Tran et al., 2018) and the model is referred to as R(2+1)D in this paper. The network is shown in Figure 3. B.

The final tested model is a pre-trained SlowFast network on the Kinetics400 dataset. Kinetics400 is a relatively recent large video classification dataset, which is focused on human actions with 400 classes. Each class has at least 400 videos taken from a different YouTube video, with each lasting around 10 seconds, for a total of 306,245 clips (Kay et al., 2017). As presented in Figure 3. C, the network is composed of two streams: a slow pathway operating at a slow frame rate and capturing spatial semantics, and a fast pathway operating at a higher frame rate and capturing motion at a finer temporal resolution (Feichtenhofer et al., 2019). Eliminating the need for optical flow, this model can be trained end-to-end. The two streams share the same ResNet50 backbone. The fast pathway is more lightweight by having $1/8^{th}$ the number of channels of every convolutional block in the slow pathway. The two pathways are fused after every convolutional block through lateral connections. Besides outperforming R(2+1)D on the Kinetics400 dataset and demonstrating faster inference time, this model was also of interest to evaluate the potential generalisation benefits of leveraging a pre-trained model on a relatively difficult and large video dataset for the melt pool use case. Further architectural and performance details can be found in Feichtenhofer et al. (2019).

## 4. IMPLEMENTATION DETAILS

All experiments were run in Google Colab Pro+ (Colab) with GPU enabled. The GPU chip type is not specified as Colab dynamically assigns them based on several usage factors. All models were built using PyTorch and its associated libraries. However, ApacheMXNet and GluonCV were used for the SlowFast model.

Three experimental settings were tested: training and validating on the original dataset, and testing on the augmented validation set (setting A); training and validating on the augmented dataset, and testing on the original validation set (setting B); and finally, training and validating on the mixed dataset (i.e., the combined original and augmented data), and individually evaluating the performance on the original and augmented validation sets (setting C). Each model was trained with different hyperparameter settings, which are listed in Table 1.

**Table 1. Model hyperparameters. LR refers to learning rate, B refers to batch size, M refers to momentum, and SGD refers to stochastic gradient descent.**

|  | **CNN1** | **CNN-LSTM** | **R(2+1)D** | **SlowFast** |
|---|---|---|---|---|
| **Input Size (C,H,W)** | Resize (1,140, 200) | Resize (3,227, 227) | Resize (3,112, 112) | Resize (1,224, 224) |
| **LR** | 0.001 | 0.01 | 0.01 | 0.001 |
| **B** | 64 | 32 | 25 | 5 |
| **Optimizer (M) [decay]** | SGD (0.9) | SGD (0.9) | Adam [5e-4] | SGD (0.9) [1e-4] |
| **Total Epochs** | 100 | 50 | 50 | 100-200 |

## 5. RESULTS AND DISCUSSION

As noted in Table 2, the first three tested models behave similarly and on par with CNN1 presented by Zhang et al. (2019). They can learn the given data but only weakly generalise. The weak generalisation can be seen by the poor performance on the unseen data (i.e., on the perturbed data when trained on the original dataset, and vice versa).

**Table 2. Accuracy results in the three experimental settings. Maximum validation and test results for each setting are emboldened. Val refers to validation, Org refers to original, and Aug refers to augmented. * indicates average results over 10 runs.**

| Dataset Setting | CNN1 | CNN-LSTM | R(2+1)D | SlowFast* |
|---|---|---|---|---|
| Setting A | Val: 85.00%, Test: 23.70% | **Val: 93.75%,** Test: 32.00% | Val: 90.17%, Test: 33.10% | Val: 65.00%, **Test: 56.70%** |
| Setting B | **Val: 98.46%,** Test: 24.40% | Val: 94.57%, Test: 27.30% | Val: 87.50%, Test: 28.00% | Val: 86.40%, **Test: 95.00%** |
| Setting C | Val: 87.69%, Test Aug: 87.62%, Test Org: 27.59% | **Val: 93.91%,** **Test Aug: 94.00%,** Test Org: 26.00% | Val: 91.85%, Test Aug: 92.00%, Test Org: 26.00% | Val: 85.25%, Test Aug: 85.26%, **Test Org: 95.00%** |

Moreover, when trained on the mixed dataset, poor validation accuracy is observed on the original data alone. This could be due to a considerably lower number of samples seen during training as compared to the augmented data. Although we were not able to reproduce the entirety of the hybrid architecture through CNN2 by Zhang et al. (2019), we do not expect the hybrid model to achieve significantly better performance than CNN1. Indeed, their results show that the combined model only improved CNN1's results by a small margin. Amongst the first three spatiotemporal classifiers, the CNN-LSTM models exhibited the overall best performance across the three settings. Despite its competitive performance, R(2+1)D proved to be computationally challenging given its depth and considerably high number of parameters.

In the case of the optical-flow-based two-stream networks (not displayed on Table 2), although we were unable to train the model with our resources, the spatial pre-trained VGG16 on ImageNet was tested by fine-tuning the last layer. The experimental setting A took a considerably long time to train (over 200 epochs) to achieve an unsatisfactory 75.73% and 75.54% training and validation accuracy, respectively. When tested on the augmented test data, a 32.67% accuracy is attained. When trained on setting C, it is observed that the model struggles to even weakly generalise to the mixed validation data despite hyperparameter tuning efforts, achieving a 71.20% training accuracy and a consistent validation accuracy fluctuating around 43.00%. Training was thus not furthered beyond 100 epochs. This demonstrates that using a pre-trained model on the static ImageNet does not aid in a model's generalisability in the case of melt pool classification.

As for the Kinetics400 pre-trained SlowFast network, the results are averaged over 10 runs due to the randomness introduced during the selection of the input data. In all three settings, a higher average validation accuracy than training accuracy was observed. Although this can present a potential source of concern with regards to overfitting, the model demonstrated relatively strong generalisation abilities on the test data across all three settings. It is worth noting that despite hyperparameter tuning (e.g., smaller learning rate of $10^{-4}$, larger batch size of 10), high instabilities during training in setting A were observed in both the training and validation accuracies. Nonetheless, the SlowFast model outperformed all other tested networks on the unseen data in setting A (i.e., on the perturbed data in this case), gaining a 56.7% average accuracy over 10 runs. In setting C, despite the fewer examples as compared to the augmented data and unlike the other models, the pre-trained SlowFast network achieved a highly competitive accuracy on the original test data. Thus, these results illustrate that leveraging spatiotemporal information in both the pre-training and the melt pool data can considerably aid in the model's robustness.

## 6. CONCLUSIONS

In conclusion, key spatiotemporal deep learning models were evaluated for print track anomaly classification. Specifically, representative architectures from recurrent spatial, two-stream, and factorized CNN categories were implemented and validated against a baseline spatial CNN. The dataset consisted of melt pool videos from a LPBF process, which were captured using a visible light camera. The generalisation abilities of the considered architectures to possible perturbations were tested using data augmentation techniques grounded in palpable process situations. It was found that the Kinetics400 pre-trained SlowFast network of the two-stream category was able to best generalise to data perturbations. For future work, the hyperparameters of the presented models can be further fine-tuned to optimal. Moreover, alternative data augmentation methods and learners with strong generalisation capacities can be tried to find an optimal architecture for the melt pool stream classification use case. Specifically, for data augmentation, deep generative approaches may be tried. As for models, more recent state-of-the-art video classifiers, in combination with renowned robust architectures, can be investigated. Finally, the architectures' compactness can be iterated for in-line inference to find an optimal accuracy-to-computational-efficiency trade-off.

## ACKNOWLEDGEMENTS

McGill Engineering Doctoral Award (MEDA) fellowship for Mutahar Safdar is acknowledged with gratitude. Mutahar Safdar also received financial support from the National Research Council of Canada (NRC INT-015-1). The authors are grateful to the Digital Research Alliance of Canada (RRG# 4294) for providing computational resources to support this research. We sincerely thank Prof. Jerry Fuh (National University of Singapore), Prof. Kunpeng Zhu (Institute of Advanced Manufacturing Technology/Institute of Intelligent Machines, Chinese Academy of Sciences), and Dr. Yingjie

Zhang (South China University of Technology) for sharing the data used in this study.